\pdfoutput=1

\documentclass[11pt]{article}

\usepackage[final]{acl}

\usepackage{times}
\usepackage{latexsym}

\usepackage[T1]{fontenc}

\usepackage[utf8]{inputenc}

\usepackage{microtype}

\usepackage{inconsolata}

\usepackage{graphicx}
\usepackage{multirow}
\usepackage[normalem]{ulem}
\usepackage{url}
\title{Bridging AI and Carbon Capture: A Dataset for LLMs in Ionic Liquids and CBE Research}

\author{Gaurab Sarkar\textsuperscript{1\thanks{Both authors contributed equally to this paper.}} and Sougata Saha\textsuperscript{2\footnotemark[1]}\\
\textsuperscript{1}State University of New York at Buffalo, Department of Chemical and Biological Engineering\\ \textsuperscript{2}Mohamed bin Zayed University of Artificial Intelligence, Department of Natural Language Processing\\
\texttt{\textsuperscript{1}gaurabsa@buffalo.edu}, \texttt{\textsuperscript{2}sougata.saha@mbzuai.ac.ae}
}

\begin{document}
\maketitle
\begin{abstract}
Large Language Models (LLMs) have demonstrated exceptional performance in general knowledge and reasoning tasks across various domains. However, their effectiveness in specialized scientific fields like Chemical and Biological Engineering (CBE) remains underexplored. Addressing this gap requires robust evaluation benchmarks that assess both knowledge and reasoning capabilities in these niche areas, which are currently lacking. To bridge this divide, we present a comprehensive empirical analysis of LLM reasoning capabilities in CBE, with a focus on Ionic Liquids (ILs) for carbon sequestration—an emerging solution for mitigating global warming. We develop and release an expert-curated dataset of 5,920 examples designed to benchmark LLMs' reasoning in this domain. The dataset incorporates varying levels of difficulty, balancing linguistic complexity and domain-specific knowledge. Using this dataset, we evaluate three open-source LLMs with fewer than 10 billion parameters. Our findings reveal that while smaller general-purpose LLMs exhibit basic knowledge of ILs, they lack the specialized reasoning skills necessary for advanced applications. Building on these results, we discuss strategies to enhance the utility of LLMs for carbon capture research, particularly using ILs. Given the significant carbon footprint of LLMs, aligning their development with IL research presents a unique opportunity to foster mutual progress in both fields and advance global efforts toward achieving carbon neutrality by 2050. Dataset link: \url{https://github.com/sougata-ub/llms_for_ionic_liquids}

\end{abstract}

\section{Introduction}

Despite notable advancements in modeling and simulation methods \cite{gunsteren1998jcp, gunsteren2018ange, frenkel2023understanding}, fundamental research in CBE continues to rely heavily on experimental results. As computational models \cite{zhao2023survey}, LLMs are predominantly advantageous in computation-intensive fields, making their precise role in enabling progress within experiment-driven domains like CBE unclear. Nonetheless, recent breakthroughs in material discovery \cite{lu2023modeling, luu2024bioinspiredllm, lu2024generative, buehler2023melm} and protein engineering \cite{jumper2021highly, liu2022presto, yu2022colgen, yu2022end, hu2022end, khare2022collagentransformer} demonstrate the potential of AI technologies to contribute meaningfully to such fields. To unlock the potential applications of LLMs in CBE, it is critical to assess their knowledge and reasoning capabilities. However, this requires robust, domain-specific evaluation benchmarks, which are currently lacking in CBE.

While evaluation frameworks exist in related fields, they predominantly rely on cloze-style tasks to assess LLMs' knowledge capacity or focus on narrow, task-specific evaluations \cite{zhao2024chemsafetybench, murakumo2023llm, zhang2024chemllm, guo2023can, bran2023chemcrow}. Such approaches are often insufficiently general and may not adequately capture the complexities of CBE. Given that LLMs have been trained on a vast corpus of publicly available online data \cite{villalobos2022will, villalobos2024will}, studies \cite{chu2025sftmemorizesrlgeneralizes} have shown that these models can easily memorize and regurgitate information during cloze-style factual assessments. This limitation provides only a superficial understanding of LLM capabilities across domains. Furthermore, the concept of knowledge extends beyond factual recall to include its application (\textit{p-knowledge}) \cite{fierro-etal-2024-defining}. Therefore, evaluating knowledge capacity alone fails to capture reasoning ability, hindering the practical deployment of LLMs, particularly in fields like CBE, where their utility remains uncertain. To address this gap, we introduce a reasoning evaluation test-bed designed to more effectively estimate LLMs' applicability in such domains.

Global warming caused by greenhouse gas emissions remains a critical challenge \cite{wang2016advances, sanz2016direct}, necessitating accelerated research into effective carbon capture solutions \cite{sheridan2018role}. Meeting the ambitious carbon-neutral target of the 2015 Paris Agreement by 2050 \cite{paris2015} requires not only reducing carbon emissions but also investing in technologies to remove CO\textsubscript{2} from the atmosphere. Among potential solutions, \textit{Ionic Liquids} (ILs) \cite{zanco2021postcombustion} stand out as promising candidates for CO\textsubscript{2} separation processes due to their non-volatile, non-toxic nature (\textit{"green solvents"}), ease of regeneration, and high CO\textsubscript{2} absorption efficiency. However, experimentation with ILs and achieving industrial scalability are resource-intensive and costly, a challenge that AI technologies like LLMs could help address. In this paper, we take a foundational step toward exploring the role of LLMs in supporting carbon capture research using ILs. Specifically, we assess the potential of general-purpose LLMs in domain-specific scenarios by constructing a test bed of 5,920 expert-curated examples, spanning varying levels of difficulty, to evaluate the factual knowledge and reasoning capabilities of these models in the context of ILs. We benchmark three open-weight LLMs—Llama 3.1-8B \cite{llama31}, Mistral-7B \cite{mistral2}, and Gemma-9B \cite{gemma2}—on this dataset. Given the absence of prior research in this area, our work represents a critical step toward identifying the potential applications of LLMs in IL research. Furthermore, leveraging LLMs for CO\textsubscript{2} capture research offers an opportunity to indirectly address concerns about their environmental impact \cite{patterson2021carbon, strubell2019energyp, faiz2023llmcarbon, li2023making, rillig2023risks} by aligning their use with climate solutions. Our contributions are as follows:
\begin{itemize}
    \item \textbf{Dataset Creation:} Using ILs for carbon capture as a use case, we construct and publicly share\footnote{Dataset available at: \url{https://github.com/sougata-ub/llms_for_ionic_liquids}} a textual entailment test bed containing 5,920 expert-curated samples designed to evaluate LLM reasoning capabilities in CBE.

    \item \textbf{Benchmarking:} We systematically benchmark three open-weight LLMs—Llama 3.1-8B, Mistral-7B, and Gemma-9B—on the test bed and share the resulting insights.

    \item \textbf{Analysis:} We discuss the implications of our results and the broader potential for LLMs to advance IL research and CO\textsubscript{2} capture technologies.
\end{itemize}

\section{Related Work}

\subsection{Ionic Liquids for Carbon Capture}
COP21 showed that amongst 196 participating countries, China, the United States, and India comprise the top three nations by share of worldwide CO\textsubscript{2} emissions. While the United States has pledged to reach “net-zero” by 2050, the deadlines set by China and India (the two most populous countries) are 2060 and 2070 respectively \cite{paris2015, guiot2016climate, dimitrov2016paris, robbins2016understand}. In an attempt to offset the rising atmospheric carbon dioxide levels, carbon sequestration has emerged as an effective field of research and the timely development of materials and methods is pivotal for the efficient capture of CO\textsubscript{2} \cite{wang2016advances, sanz2016direct}. Ionic Liquids have presented themselves as an excellent solution for CO\textsubscript{2} capture due to their environmentally friendly nature \cite{blanchard1999green, blanchard2001high, perez2003solubility, anthony2002solubilities, zeng2017ionic, husson2003solubilities, aghaie2018systematic, ramdin2012state}. Thorough experimentation, with ILs, to provide a practical solution is time-conducive and entails high cost \cite{sheridan2018role, maginn2009molecular}. In that regard, various machine learning methods have found use to alleviate dependence on experiments \cite{cao2018using, baskin2022benchmarking, dhakal2022generalized, feng2022estimation, paduszynski2016silico}.

\subsection{LLMs for Scientific Research}

Recently, LLMs \cite{brown2020language, chowdhery2023palm, taylor2022galactica, openai2024gpt4technicalreport} have gained significant popularity with a wide range of possibilities \cite{ge2024openagi, bubeck2023sparks, nadkarni2021scientific, beltagy2019scibert, schick2024toolformer, buehler2023generative, luu2023generative, mialon2023augmented, wei2023chainofthought}, and the integration of these transformer-based models into the fields of materials science and discovery has yielded tremendous results. Leveraging the abilities of LLMs has been beneficial in various downstream tasks such as protein design and folding \cite{jumper2021highly, liu2022presto, yu2022colgen, yu2022end, hu2022end, khare2022collagentransformer}, material discovery \cite{lu2023modeling, luu2024bioinspiredllm, lu2024generative, buehler2023melm}, educational tasks \cite{lim2023generative, milano2023large, inguva2021introducing} and chemistry-related tasks \cite{castro2023large, white2023future, jablonka202314}. The reliability of LLMs is still a massive topic of discussion, and their accuracy is often determined by the size and complexity of the model. Despite their promises, present pitfalls include the issues of hallucinations and fact recall, which warrants a careful validation of the model's output and its eventual ramifications \cite{hu2023deep, azamfirei2023large, kandpal2023large, varshneystitch, ji2023survey, mckenna2023sources, harrer2023attention}. Invariably, training and using such networks comes at a huge environmental cost, largely in terms of carbon emissions \cite{li2023making, patterson2021carbon, strubell2019energyp, faiz2023llmcarbon, rillig2023risks}.

The power of LLMs can aid carbon capture by helping researchers with their advances to address the growing problem of global warming and offset the model’s carbon footprint to reach the end goal of 'net-zero' carbon emissions.

\section{A Practical Test for Knowledge}
Although there are several standard definitions of knowledge in Philosophy \cite{sartwell1992knowledge, nozick2016knowledge, williamson2005knowledge, zagzebski2017knowledge, austin1961other}, the most prevalent ones for non-human entities like LLMs are \textit{tb} and \textit{p-knowledge} \cite{fierro-etal-2024-defining}. Most knowledge probing tasks test for \textit{tb-knowledge}, where the model passes the test if it can recall an answer. For example, probing for factual questions like "What is the capital of Germany?" Such tests are weak estimates of knowledge and hold little pragmatic significance, especially in domains like CBE, where the intended use of LLMs is still unclear. LLMs as reasoners can be of better practical use in such domains. Although some methods estimate the model's uncertainty \cite{huang2024survey, huang2023look, ye2024benchmarking, geng2023survey}, they still pertain to \textit{tb-knowledge}. However, a more complete measure of knowledge is \textit{p-knowledge}, which tests a model's capability to use knowledge in practical tasks. For example, sociodemographic prompting \cite{saha2025reading, pandey-etal-2025-culturally, li2024culture, alkhamissi-etal-2024-investigating, nadeem-etal-2021-stereoset,nangia-etal-2020-crows, wan-etal-2023-personalized, jha-etal-2023-seegull, li2024culturegenrevealingglobalcultural, cao-etal-2023-assessing, tanmay2023probingmoraldevelopmentlarge, rao-etal-2023-ethical} such as "What would a German find difficult to understand from a text X?" necessitates a model to reason from a group's perspective, which requires prerequisite knowledge. Motivated to create stronger test beds, we set up an \textit{entailment task} to benchmark LLMs' factual capacity in CBE, where the model is provided a claim and a list of propositions and is tasked with determining all propositions that entail the claim or none. Thus, testing the model's reasoning capabilities in a practical setting is warranted.

\subsection{Argument Structures}
A claim constitutes one or more facts (propositions), where some are evident (explicit) from the text, and some are assumed (implicit) to be known by the reader (enthymemes) \cite{walton1996argument, besnard2008elements, walton2008argumentation, bitzer2020aristotle}. Within a field (such as CBE), the degree of knowledge of the assumed propositions is subjective and varies by person, which impacts the understanding of the claim. For example, the claim "Ionic Liquids are low-melting, non-volatile salts which categorize them within the green solvents category" explicitly informs that (i) Ionic Liquids are low-melting, non-volatile salts. (ii) Ionic Liquids are categorized as green solvents. It also entails that low-melting and non-volatile salts are green solvents, which might be unknown (or partially known) to someone from CBE\footnote{This is different from general knowledge. For example, understanding the claim also requires knowledge of "low-melting, non-volatile salts" and "green solvents", which is an assumed prerequisite for a domain expert.}. The degree of knowledge about the implicit assumption is subjective and varies within the domain\footnote{We are only interested in domain-specific knowledge. An outsider might not possess such knowledge.}. We aim to test this domain-specific knowledge in LLMs via an entailment task.

\subsection{The Entailment Task}
\label{hypothesis}

Hypothesizing that \textbf {knowledgeable agents should perform consistently, irrespective of the adversaries}, we create an \textit{entailment task} with the following setups to benchmark LLMs' reasoning capacity, where the model is provided a claim and a list of propositions and tasked to determine all propositions that entail the claim, if applicable. 

\noindent
\textbf{1. Change the number of adversaries:} (i) Keeping the number of entailing propositions constant for a claim, the number of non-entailing propositions should not affect the model's entailment performance. (ii) When provided with only non-entailing options and an additional \textit{"none of the above"} option, a consistent agent should always choose the \textit{"none"} option. A drop in performance indicates a lack of knowledge and supposedly more reliance on linguistic cues for entailment.

\noindent
\textbf{2. Introduce linguistic perturbations:} A knowledgeable agent should be invariant to paraphrased options. Failure to do so indicates reliance on linguistic cues instead of factual cues for entailment.

\noindent
\textbf{2. Apply common sense:} Knowledgeable agents should not be derailed by incorrect facts that can be discerned by common sense.


\begin{figure*}[th!]
    \centering
    \includegraphics[width=\linewidth]{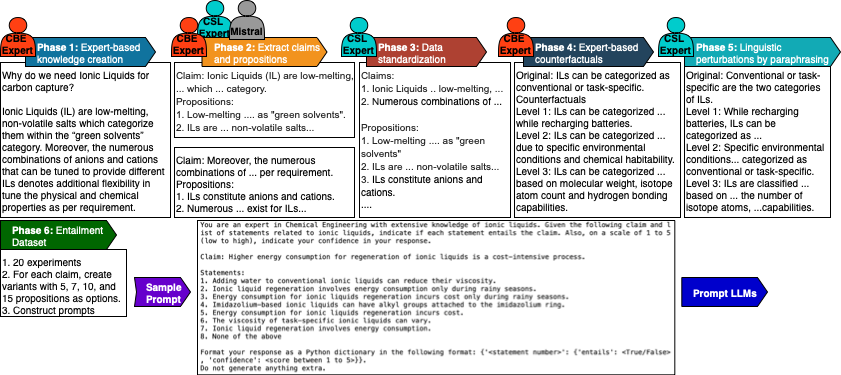}
    \caption{Dataset Creation Pipeline}
    \label{fig:dataset_creation}
\end{figure*}

\subsection{Dataset Creation}
The dataset is created in multiple phases, employing two expert annotators, one with a background in CBE and another from Computer Science and Linguistics (CSL). The CBE expert has domain knowledge of ILs for carbon capture, while the CSL expert is generally unaware of the domain. Figure \ref{fig:dataset_creation} illustrates the data creation pipeline with an actual example. We detail the pipeline below:

\noindent
\textbf{Phase 1} encompassed knowledge creation, where the CBE expert constructed paragraphs capturing the different aspects of carbon capture using ionic liquids. The aspects encompassed the need for carbon capture, ionic liquids, their physical and chemical characteristics, and their advantages. Next, the annotator extracted claims from the paragraphs, which are sentences containing salient knowledge pertaining to ionic liquids for carbon capture, yielding 74 in total.

\noindent
\textbf{Phase 2} encompassed identifying the explicit and implicit propositions from each claim and implemented in two stages: (i) \textbf{LLM-based annotation:} We prompted Mistral-7B-Instruct-v0.2 \cite{mistral2} to identify the explicit and implicit propositions from a claim. As depicted in Figure \ref{fig:mistral_assumption_extraction_prompt} (Appendix \ref{sec:appendix}), the prompt comprised a short task description and three examples of how to perform the task, followed by the actual claim for annotation. (ii) \textbf{Expert evaluations:} The CBE expert extensively evaluated the model response by editing, deleting, or unchanging each model-identified proposition. Additionally, for each claim, the expert added propositions that were missed by the model (if any). Overall, 48 (65\%) of the 74 LLM-based annotations were deemed correct by the expert and were unmodified, yielding 164 propositions across 74 claims.


\noindent
\textbf{Phase 3} encompassed data standardization. The propositions, being fundamental pieces of knowledge, are universal. Hence, in this phase, we standardized the propositions across all claims. Using sentence transformers \cite{reimers-2019-sentence-bert}, we clustered the propositions by their embedding cosine-similarity\footnote{We used the 'all-MiniLM-L6-v2' model for computing embeddings.} and computationally marked propositions belonging to the same cluster as equivalent. The CBE expert evaluated the clustering results, which were accurate in only 28\% of cases. The expert annotated and rectified the incorrect cluster assignments, yielding 125 universal propositions across all 74 claims.


\begin{table*}[t!]
\centering
\resizebox{\linewidth}{!}{%
\begin{tabular}{|c|l|c|l|cc|cc|ccc|ccc|}
\hline
\multirow{2}{*}{\textbf{Group}} &
  \multicolumn{1}{c|}{\multirow{2}{*}{\textbf{Description}}} &
  \multirow{2}{*}{\textbf{Id}} &
  \multicolumn{1}{c|}{\multirow{2}{*}{\textbf{Experiment}}} &
  \multicolumn{2}{c|}{\textbf{Correct options}} &
  \multicolumn{2}{c|}{\textbf{Incorrect options}} &
  \multicolumn{3}{c|}{\textbf{Median F1}} &
  \multicolumn{3}{c|}{\textbf{Std Dev F1}} \\ \cline{5-14} 
 &
  \multicolumn{1}{c|}{} &
   &
  \multicolumn{1}{c|}{} &
  \multicolumn{1}{c|}{\textbf{Present}} &
  \textbf{\begin{tabular}[c]{@{}c@{}}Para-\\ phrased\end{tabular}} &
  \multicolumn{1}{c|}{\textbf{Difficulty}} &
  \textbf{\begin{tabular}[c]{@{}c@{}}Para-\\ phrased\end{tabular}} &
  \multicolumn{1}{c|}{\textbf{Gemma}} &
  \multicolumn{1}{c|}{\textbf{Llama}} &
  \textbf{Mistral} &
  \multicolumn{1}{c|}{\textbf{Gemma}} &
  \multicolumn{1}{c|}{\textbf{Llama}} &
  \textbf{Mistral} \\ \hline
\multirow{2}{*}{0} &
  \multirow{2}{*}{Baselines} &
  1 &
  orig+random &
  \multicolumn{1}{c|}{\multirow{2}{*}{Yes}} &
  No &
  \multicolumn{1}{c|}{\multirow{2}{*}{Random}} &
  \multirow{2}{*}{No} &
  \multicolumn{1}{c|}{49.0} &
  \multicolumn{1}{c|}{\textbf{66.0}} &
  55.0 &
  \multicolumn{1}{c|}{21.7} &
  \multicolumn{1}{c|}{\textbf{16.1}} &
  18.9 \\ \cline{3-4} \cline{6-6} \cline{9-14} 
 &
   &
  2 &
  para+random &
  \multicolumn{1}{c|}{} &
  Yes &
  \multicolumn{1}{c|}{} &
   &
  \multicolumn{1}{c|}{49.5} &
  \multicolumn{1}{c|}{\textbf{63.0}} &
  57.0 &
  \multicolumn{1}{c|}{21.8} &
  \multicolumn{1}{c|}{\textbf{14.0}} &
  18.5 \\ \hline
\multirow{6}{*}{1} &
  \multirow{6}{*}{\begin{tabular}[c]{@{}l@{}}Only\\ providing\\ incorrect\\ options\end{tabular}} &
  3 &
  none+level1 &
  \multicolumn{1}{c|}{\multirow{6}{*}{No}} &
  \multirow{6}{*}{No} &
  \multicolumn{1}{c|}{Level 1} &
  \multirow{3}{*}{No} &
  \multicolumn{1}{c|}{9.0} &
  \multicolumn{1}{c|}{\textbf{30.0}} &
  1.5 &
  \multicolumn{1}{c|}{5.5} &
  \multicolumn{1}{c|}{6.8} &
  \textbf{3.1} \\
 &
   &
  4 &
  none+level2 &
  \multicolumn{1}{c|}{} &
   &
  \multicolumn{1}{c|}{Level 2} &
   &
  \multicolumn{1}{c|}{2.0} &
  \multicolumn{1}{c|}{\textbf{29.0}} &
  0.0 &
  \multicolumn{1}{c|}{4.9} &
  \multicolumn{1}{c|}{12.3} &
  \textbf{1.0} \\
 &
   &
  5 &
  none+level3 &
  \multicolumn{1}{c|}{} &
   &
  \multicolumn{1}{c|}{Level 3} &
   &
  \multicolumn{1}{c|}{3.0} &
  \multicolumn{1}{c|}{\textbf{21.5}} &
  0.0 &
  \multicolumn{1}{c|}{4.3} &
  \multicolumn{1}{c|}{8.6} &
  \textbf{0.5} \\
 &
   &
  6 &
  none+level1-para &
  \multicolumn{1}{c|}{} &
   &
  \multicolumn{1}{c|}{Level 1} &
  \multirow{3}{*}{Yes} &
  \multicolumn{1}{c|}{3.0} &
  \multicolumn{1}{c|}{\textbf{17.5}} &
  0.0 &
  \multicolumn{1}{c|}{1.7} &
  \multicolumn{1}{c|}{7.3} &
  \textbf{0.5} \\
 &
   &
  7 &
  none+level2-para &
  \multicolumn{1}{c|}{} &
   &
  \multicolumn{1}{c|}{Level 2} &
   &
  \multicolumn{1}{c|}{0.0} &
  \multicolumn{1}{c|}{\textbf{17.5}} &
  0.0 &
  \multicolumn{1}{c|}{2.0} &
  \multicolumn{1}{c|}{6.4} &
  \textbf{0.5} \\
 &
   &
  8 &
  none+level3-para &
  \multicolumn{1}{c|}{} &
   &
  \multicolumn{1}{c|}{Level 3} &
   &
  \multicolumn{1}{c|}{1.0} &
  \multicolumn{1}{c|}{\textbf{18.5}} &
  0.0 &
  \multicolumn{1}{c|}{5.3} &
  \multicolumn{1}{c|}{2.4} &
  \textbf{0.0} \\ \hline
\multirow{3}{*}{2} &
  \multirow{3}{*}{\begin{tabular}[c]{@{}l@{}}Difficulty le-\\vel of incor-\\rect options\end{tabular}} &
  9 &
  orig+level1 &
  \multicolumn{1}{c|}{\multirow{3}{*}{Yes}} &
  \multirow{3}{*}{No} &
  \multicolumn{1}{c|}{Level 1} &
  \multirow{3}{*}{No} &
  \multicolumn{1}{c|}{35.0} &
  \multicolumn{1}{c|}{\textbf{73.5}} &
  62.5 &
  \multicolumn{1}{c|}{26.6} &
  \multicolumn{1}{c|}{\textbf{14.6}} &
  19.6 \\
 &
   &
  10 &
  orig+level2 &
  \multicolumn{1}{c|}{} &
   &
  \multicolumn{1}{c|}{Level 2} &
   &
  \multicolumn{1}{c|}{32.0} &
  \multicolumn{1}{c|}{\textbf{68.5}} &
  60.5 &
  \multicolumn{1}{c|}{19.9} &
  \multicolumn{1}{c|}{\textbf{12.3}} &
  18.7 \\
 &
   &
  11 &
  orig+level3 &
  \multicolumn{1}{c|}{} &
   &
  \multicolumn{1}{c|}{Level 3} &
   &
  \multicolumn{1}{c|}{29.5} &
  \multicolumn{1}{c|}{\textbf{67.5}} &
  58.0 &
  \multicolumn{1}{c|}{18.0} &
  \multicolumn{1}{c|}{\textbf{12.1}} &
  16.9 \\ \hline
\multirow{3}{*}{3} &
  \multirow{3}{*}{\begin{tabular}[c]{@{}l@{}}Paraphrasing\\ the correct\\ options\end{tabular}} &
  12 &
  para+level1 &
  \multicolumn{1}{c|}{\multirow{3}{*}{Yes}} &
  \multirow{3}{*}{Yes} &
  \multicolumn{1}{c|}{Level 1} &
  \multirow{3}{*}{No} &
  \multicolumn{1}{c|}{40.5} &
  \multicolumn{1}{c|}{\textbf{66.5}} &
  62.0 &
  \multicolumn{1}{c|}{24.1} &
  \multicolumn{1}{c|}{\textbf{14.4}} &
  20.1 \\
 &
   &
  13 &
  para+level2 &
  \multicolumn{1}{c|}{} &
   &
  \multicolumn{1}{c|}{Level 2} &
   &
  \multicolumn{1}{c|}{34.5} &
  \multicolumn{1}{c|}{\textbf{66.0}} &
  60.0 &
  \multicolumn{1}{c|}{23.3} &
  \multicolumn{1}{c|}{\textbf{13.5}} &
  18.1 \\
 &
   &
  14 &
  para+level3 &
  \multicolumn{1}{c|}{} &
   &
  \multicolumn{1}{c|}{Level 3} &
   &
  \multicolumn{1}{c|}{31.5} &
  \multicolumn{1}{c|}{\textbf{64.0}} &
  60.0 &
  \multicolumn{1}{c|}{18.8} &
  \multicolumn{1}{c|}{\textbf{13.0}} &
  17.0 \\ \hline
\multirow{3}{*}{4} &
  \multirow{3}{*}{\begin{tabular}[c]{@{}l@{}}Paraphrasing\\ the incorrect\\ options\end{tabular}} &
  15 &
  orig+level1-para &
  \multicolumn{1}{c|}{\multirow{3}{*}{Yes}} &
  \multirow{3}{*}{No} &
  \multicolumn{1}{c|}{Level 1} &
  \multirow{3}{*}{Yes} &
  \multicolumn{1}{c|}{39.5} &
  \multicolumn{1}{c|}{\textbf{66.0}} &
  57.0 &
  \multicolumn{1}{c|}{22.1} &
  \multicolumn{1}{c|}{\textbf{11.7}} &
  17.3 \\
 &
   &
  16 &
  orig+level2-para &
  \multicolumn{1}{c|}{} &
   &
  \multicolumn{1}{c|}{Level 2} &
   &
  \multicolumn{1}{c|}{35.5} &
  \multicolumn{1}{c|}{\textbf{64.5}} &
  57.5 &
  \multicolumn{1}{c|}{17.0} &
  \multicolumn{1}{c|}{\textbf{11.1}} &
  17.1 \\
 &
   &
  17 &
  orig+level3-para &
  \multicolumn{1}{c|}{} &
   &
  \multicolumn{1}{c|}{Level 3} &
   &
  \multicolumn{1}{c|}{34.0} &
  \multicolumn{1}{c|}{\textbf{63.0}} &
  58.5 &
  \multicolumn{1}{c|}{13.5} &
  \multicolumn{1}{c|}{\textbf{10.5}} &
  16.6 \\ \hline
\multirow{3}{*}{5} &
  \multirow{3}{*}{\begin{tabular}[c]{@{}l@{}}Paraphrasing\\ all options\end{tabular}} &
  18 &
  para+level1-para &
  \multicolumn{1}{c|}{\multirow{3}{*}{Yes}} &
  \multirow{3}{*}{Yes} &
  \multicolumn{1}{c|}{Level 1} &
  \multirow{3}{*}{Yes} &
  \multicolumn{1}{c|}{33.5} &
  \multicolumn{1}{c|}{\textbf{63.5}} &
  58.0 &
  \multicolumn{1}{c|}{\textbf{10.7}} &
  \multicolumn{1}{c|}{11.2} &
  16.9 \\
 &
   &
  19 &
  para+level2-para &
  \multicolumn{1}{c|}{} &
   &
  \multicolumn{1}{c|}{Level 2} &
   &
  \multicolumn{1}{c|}{35.5} &
  \multicolumn{1}{c|}{\textbf{64.5}} &
  59.5 &
  \multicolumn{1}{c|}{15.6} &
  \multicolumn{1}{c|}{\textbf{11.8}} &
  17.6 \\
 &
   &
  20 &
  para+level3-para &
  \multicolumn{1}{c|}{} &
   &
  \multicolumn{1}{c|}{Level 3} &
   &
  \multicolumn{1}{c|}{36.5} &
  \multicolumn{1}{c|}{\textbf{60.5}} &
  58.5 &
  \multicolumn{1}{c|}{11.9} &
  \multicolumn{1}{c|}{\textbf{10.8}} &
  17.8 \\ \hline
\end{tabular}%
}
\caption{Definitions of experiments and aggregated model results (median F1 and standard deviation) across experiments with 5, 7, 10, and 15 options. The best scores are highlighted in bold.}
\label{tab:f1-scores}
\end{table*}

\noindent
\textbf{Phase 4} involved constructing false variants of the propositions at three difficulty levels: (i) \textbf{Low:} Invalid version of a proposition, and can be discerned using common sense reasoning. For example, the proposition \textit{"Ionic liquids can be categorized as conventional or task-specific"} was augmented to \textit{"Ionic liquids can be categorized as conventional or task-specific only while recharging batteries."} (ii) \textbf{Medium:} Invalid version of a proposition that might need a mix of common sense and knowledge of science for discerning. For example, \textit{"Ionic liquids can be categorized as conventional or task-specific due to specific environmental conditions and chemical habitability."} (iii) \textbf{High:} Determining invalidity requires considerable knowledge about ILs. For example, \textit{"Ionic liquids can be categorized as conventional or task-specific based on molecular weight, isotope atom count, and hydrogen bonding capabilities."} All variants were manually constructed by the CBE expert and evaluated by the CSL expert, who does not know ILs. The CSL expert evaluated 60 random propositions (15 original and 15 from each level of difficulty) by determining if the proposition was correct or assigning a level of difficulty if they thought it was incorrect. Comparing their response with the original labels, the expert attained an F1 score of 67\% in discerning factual correctness. For the incorrect propositions, the expert attained F1 scores of 80\%, 15\%, and 42\% for levels 1, 2, and 3, indicating the difficulty of the options for a non-expert.

\noindent
In \textbf{phase 5}, we introduced linguistic variations in the original and all three incorrect variants of each proposition by paraphrasing. We prompted the Llama-3.1-8B instruction-tuned variant \cite{llama31} using the prompt \textit{"Paraphrase the following text without changing the meaning of the text. Text: <text>"} and resorted to greedy decoding for paraphrasing.

\noindent
Using the 74 claims, the 125 original propositions, and their incorrect and paraphrased variants, we constructed the test set for the entailment task in \textbf{phase 6}. For each claim, we created variants with 5, 7, 10, and 15 propositions as options. Listed in Table \ref{tab:f1-scores}, we constructed 20 experiments using different permutations of the original and paraphrased versions of the correct and incorrect propositions, yielding a dataset of 5,920 examples. On average, the claims contain 14 words, the original propositions contain 12, and the incorrect propositions contain 17.

\subsection{Experiments}
Listed in Table \ref{tab:f1-scores}, we group the 20 experiments into five groups and test our three hypotheses in Section \ref{hypothesis}. Comprising two experiments, Group 0 serves as the baseline. By only providing incorrect propositions with their difficulty and stylistic variations, the experiments in Group 1 test the model's knowledgeability by measuring the propensity of selecting the "none" option. Group 2 quantifies the effect of varying the difficulty levels of the incorrect options while keeping the original propositions unchanged. Group 3 perturbs the original proposition by paraphrasing and measures the impact of changing the incorrect option difficulty levels. Groups 4 and 5 measure a model's invariance to linguistic variations and the difficulty levels of the incorrect options. For all groups, we experiment with 5, 7, 10, and 15 propositions as options to test for the model's capability of being invariant to additional incorrect options. Except for group 1, the number of correct options varies from 1 to 5. As depicted in Figure \ref{fig:dataset_creation} (Phase 6), we construct prompts from each example and probe the Llama-3.1-8B-Instruct, Mistral-7B-Instruct-v0.3, and gemma-2-9b-it, setting the temperature to 0. We process each model response and use Llama-3.1-8B-Instruct to rectify ill-formatted outputs. Figures \ref{fig:sample_prompt_entailment} and \ref{fig:sample_prompt_correction} (Appendix \ref{sec:appendix}) illustrate the prompts for the entailment task and correcting the ill-formatted LLM responses.

\begin{figure*}[th!]
    \centering
    \includegraphics[width=\linewidth]{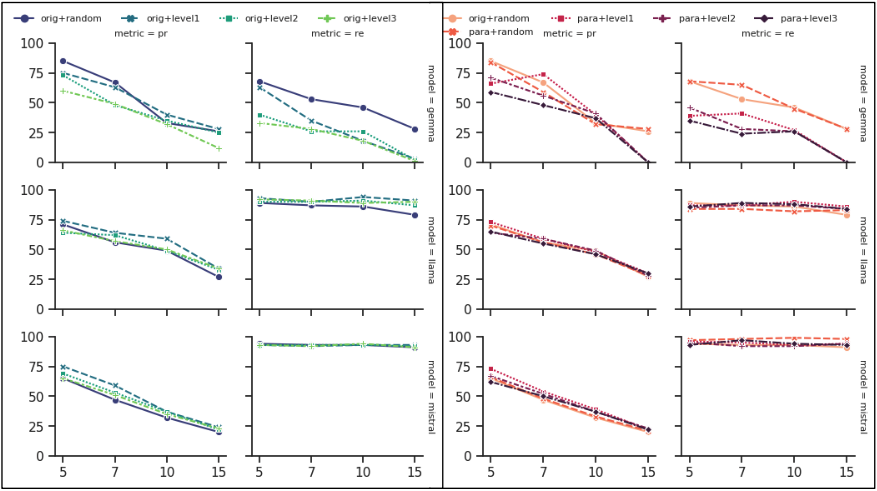}
    \caption{Model-wise precision and recall for experiments in Group 2 (left) and Group 3 (right).}
    \label{fig:exp_1_2_pr_rl}
\end{figure*}

\begin{figure*}[th!]
    \centering
    \includegraphics[width=\linewidth]{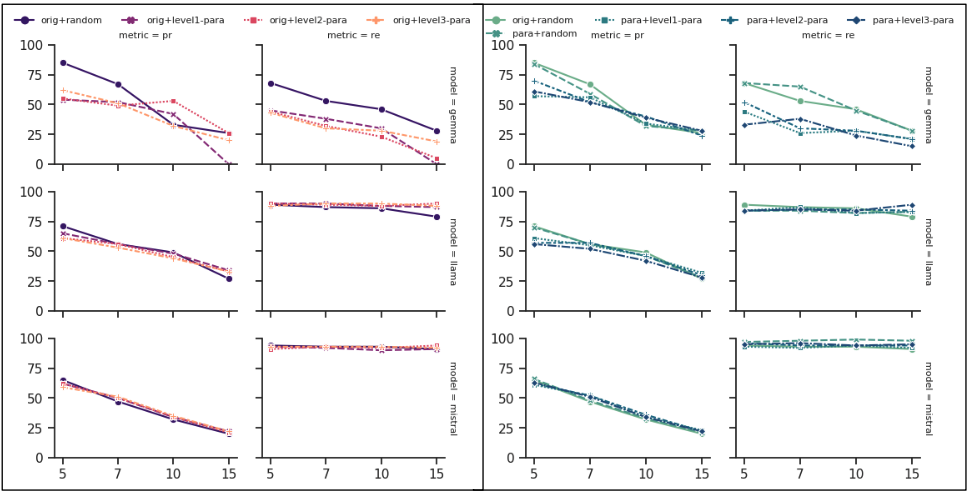}
    \caption{Model-wise precision and recall for experiments in Group 4 (left) and Group 5 (right).}
    \label{fig:exp_3_4_pr_rl}
\end{figure*}

\begin{figure*}[th!]
    \centering
    \includegraphics[width=0.75\linewidth]{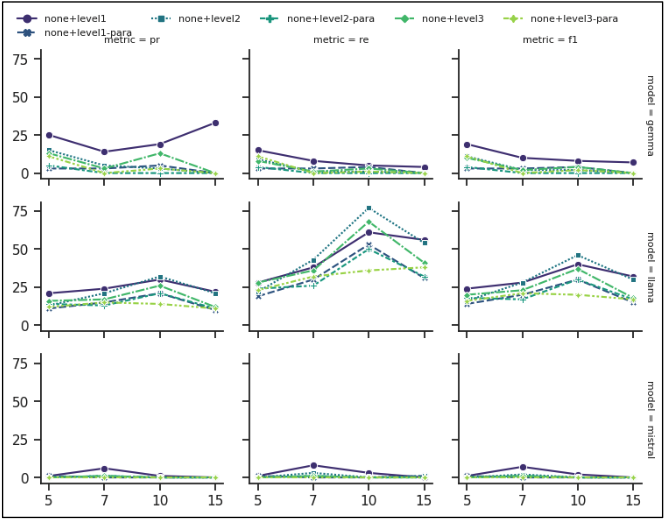}
    \caption{Model-wise precision, recall, and F1 for experiments in comparison suite 5.}
    \label{fig:only_incorrect_pr_rl_f1}
\end{figure*}

\section{Results and Observations}

Table \ref{tab:f1-scores} shares the model-wise median F1 score and the standard deviation across all options (5, 7, 10, and 15). Figures \ref{fig:exp_1_2_pr_rl}, \ref{fig:exp_3_4_pr_rl}, and \ref{fig:only_incorrect_pr_rl_f1} plot the precision and recall scores for all groups of experiments. The baseline results (Group 0) in Table \ref{tab:f1-scores} indicate that \uline{LLMs are knowledgeable about ionic liquids and carbon capture}. Llama performs the best, followed by Mistral and Gemma. However, paraphrasing the original propositions (Id 2) reduces Llama's performance, which contrasts with Mistral and Gemma, where the performance increases. This effect of stylistic perturbations on the model results shows a tendency to rely on linguistic cues. 

\noindent
\textbf{Effect of the number of incorrect options}\\
We observe a correlation between model performance and the number of incorrect options in Figures \ref{fig:exp_1_2_pr_rl}, \ref{fig:exp_3_4_pr_rl}, and \ref{fig:only_incorrect_pr_rl_f1}. The precision scores for all models drop with more incorrect options, indicating an adverse effect of the number of adversaries on their reasoning capabilities. For Llama and Mistral, the recall scores remain mostly consistent, but drop for Gemma. Nonetheless, as depicted in Table \ref{tab:f1-scores}, the standard deviation of Llama is the lowest, followed by Mistral and Gemma. For Llama and Mistral, this decline in precision but constant recall scores indicates a propensity to make more predictions as the number of options increases without changing the prediction for the correct propositions. On the contrary, increasing the number of adversaries causes Gemma to change the prediction for the correct propositions, \uline{indicating an unreliability of utilizing facts for reasoning}.

\noindent
\textbf{Effect of the difficulty of incorrect options}\\
Comparing experiment Id 1 with Group 2 and Id 2 with Group 3 in Table \ref{tab:f1-scores} and Figure \ref{fig:exp_1_2_pr_rl}, we observe that increasing the difficulty level of the adversarial facts hampers the model performance for Llama and Mistra, which is the opposite for Gemma. The comparisons indicate that the \uline{experiments comprising random adversaries (Orig/para+random) are more challenging test beds than the difficulty-controlled adversaries}, especially for Llama and Mistral. We hypothesize that since we gradually balance between common sense and domain-specific knowledge across three difficulty levels, higher performance in level 1 can be due to the model's capability of common sense reasoning, which decreases as the difficulty increases, requiring more domain-specific knowledge. However, using random adversaries presents less scope for common-sense reasoning and requires domain-knowledge-based reasoning for entailment resolution. Gemma, on the other hand, is more reliant on syntactic cues than reasoning. Hence, it falters when provided with factually incorrect yet syntactically similar options to the claim. This is also evident from Gemma's decreasing recall scores in Figure \ref{fig:exp_1_2_pr_rl}, compared to Llama and Mistral, which are more consistent.

\noindent
\textbf{Effect of only incorrect propositions as options}\\
Compared to the baseline (Group 0) in Table \ref{tab:f1-scores}, in Group 1, the performance of all models drastically reduces when presented with only incorrect facts and a "none" option to choose from. Mistral and Gemma perform worse than Llama, with median F1 scores < 10 for all experiments and near zero for some. All models perform worse with paraphrased incorrect options. 
Figure \ref{fig:only_incorrect_pr_rl_f1} plots the \textbf{pr}ecision, \textbf{re}call, and \textbf{f1} scores for Group 1 experiments. Interestingly, for all three models, sometimes the precision increases with higher options in some experiments. For Gemma, the precision scores increase while the recall decreases with an increase in incorrect choices. On the contrary, for Llama and Mistral, the precision and recall scores increase for some experiments. For Llama, presenting 7 and 10 options yields higher F1 scores for most experiments compared to 5 options. Mistral yields higher F1 scores when prompted with 7 choices compared to other options. We hypothesize that for Llama and Mistral, increasing the choices provides more inter-option reasoning opportunities, resulting in higher F1 scores. We also think the position of the "none" option in the prompt might be a confounding variable, which we leave for future work. Nonetheless, when only presented with incorrect facts and a "none" option, the drastic reduction in performance for all models indicates that \uline{although LLMs contain facts about ionic liquids, they can't reliably utilize and reason with them for complex tasks}.

\noindent
\textbf{Effect of paraphrasing}\\
Comparing Groups 2 and 3 in Table \ref{tab:f1-scores}, although paraphrasing the correct options reduces the F1 score across all difficulty levels for Llama and Mistral, paraphrasing the incorrect options in Group 4 has a higher diminishing effect on the model performance than Group 2, which is the opposite for Gemma. We hypothesize that this might be due to Gemma's reliance on linguistic cues for entailment compared to Llama and Mistral, where Gemma relies more on syntactic similarity than semantics.

Comparing Groups 3 and 5, paraphrasing the incorrect options reduces the F1 score across all difficulty levels for Llama and Mistral, which is the opposite for Gemma, except for experiment 15. Comparing Groups 4 and 5, paraphrasing the correct options reduces the F1 score for Llama across all difficulty levels. On the contrary, the F1 score increases or remains the same for Gemma and Mistral, except for experiment 18. We hypothesize that since the correct and incorrect options share syntactic similarities, they get equally transformed while paraphrasing, causing their paraphrased versions to maintain syntactic similarity, which weaker reasoning models like Gemma exploit. We leave the testing out of this hypothesis as future work.

Overall, Llama performs best across all experiments, followed by Mistral and Gemma. Our results indicate that \uline{although LLMs possess knowledge of ionic liquids and carbon capture, their domain-specific reasoning capabilities are limited}. The performance drop in Group 1 experiments is drastic for all models and sometimes near zero for Mistral and Gemma, which questions their reasoning capabilities.

\section{Discussion}
Our experiments indicate that smaller LLMs struggle to coherently reason within the domain-specific constraints and choose non-probable options in the entailment task. This is likely because LLMs are general-purpose and not geared to niche domains such as ILs. We propose that LLMs should be fine-tuned for CBE using curated datasets. Pre-training the models on domain-specific data, fine-tuning using PEFT \cite{peft} methods like LoRA \cite{lora}, or in-context learning and efficient methods such as RAG \cite{rag1, rag2} should help impart the domain-specific knowledge and constraints, which requires collaborative advancements in the intersection of LLMs and CBE. Such domain-specific LLMs can scale IL research by assisting researchers in the bottlenecked areas of data analysis, experiment design, and property predictions. Furthermore, they can serve as educational guides to researchers willing to gain familiarity with the field. This work should be a valuable resource for researchers eager to evaluate LLMs for varied fields and collaboratively help attain the sustainability goals of the UN\footnote{https://sdgs.un.org/goals}.

\section{Conclusion}

Global warming remains a pressing challenge, necessitating scalable and interdisciplinary solutions such as carbon capture. To address this need, we propose leveraging LLMs to support research on Ionic Liquids, a promising avenue for carbon capture. As a foundational step, we construct and publicly share an expert-curated dataset designed to evaluate LLMs' knowledge and reasoning capabilities within the specialized domain of Ionic Liquids. Our benchmarking of three open-weight LLMs—Llama, Gemma, and Mistral—reveals that while general-purpose models, particularly Llama, demonstrate a strong grasp of Ionic Liquid-related knowledge, they fall short in domain-specific reasoning tasks. Building on these findings, we outline potential pathways for LLMs to advance Ionic Liquid research, including their use as agents in simulations, reasoners for material discovery and design, and educational tools to help researchers familiarize themselves with the field. Moreover, optimizing LLMs for climate research not only advances carbon capture efforts but also offers a dual benefit by mitigating the models' own carbon footprint. This alignment between AI innovation and environmental goals supports the broader aim of achieving carbon neutrality by 2050.


\section*{Limitations}
This study has some notable limitations. Firstly, we only evaluate three open-weight models with less than 10B parameters for their knowledge and reasoning ability with ILs. Although extraneous experiments with larger and open-API models indicate a similar trend, they are not quantified and non-generalizable. Secondly, our entailment test set is not an exhaustive resource for IL research. It contains limited facts and only tests reasoning capabilities through entailment. We need more diverse datasets that probe the reasoning capabilities of LLMs from multiple aspects. Thirdly, we do not experiment with fine-tuning the models on our dataset and measure their impact on reasoning, which we intend as future work. Also, our work is limited to two expert evaluators and might benefit from multiple experts. Despite these limitations, our research takes a foundational step in the interdisciplinary field of LLMs for ionic liquid research, which is very nascent.

\section*{Ethics Statement}
We confirm that all conducted experiments are solely for academic purposes and adhere to ethical standards. The expert evaluators were appropriately compensated for their tasks, following all administrative and regulatory policies. The shared dataset strictly pertains to ionic liquids. It does not contain potentially explicit and sensitive content that might exhibit bias, be hurtful, or offend anyone.

\bibliography{custom}

\appendix

\section{Appendix}
\label{sec:appendix}

\begin{figure*}[th!]
    \centering
    \includegraphics[width=\linewidth]{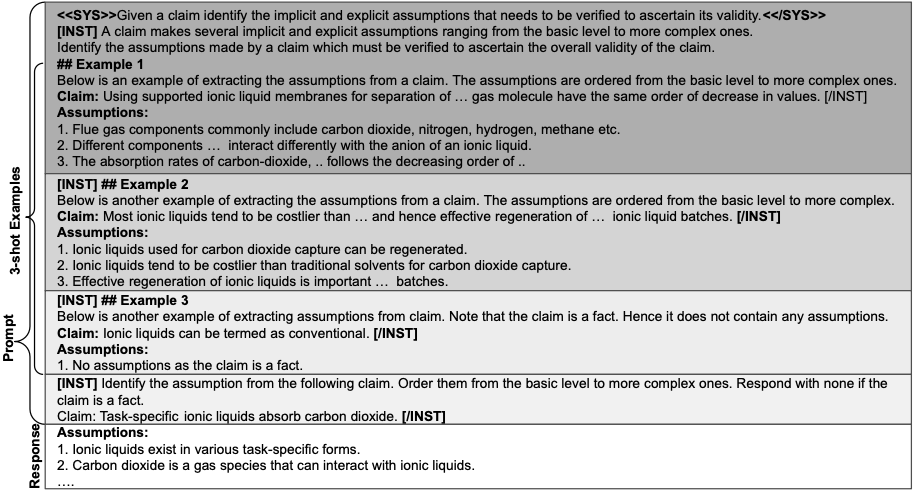}
    \caption{Mistral 3-shot prompt to automatically extract and generate the missing assumptions from claims.}
    \label{fig:mistral_assumption_extraction_prompt}
\end{figure*}

\begin{figure*}[th!]
    \centering
    \includegraphics[width=\linewidth]{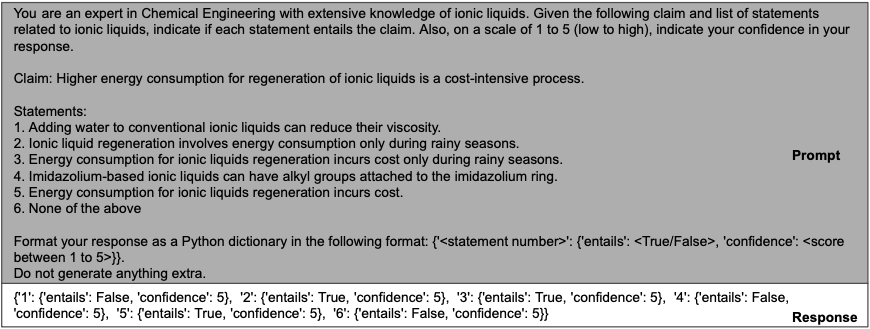}
    \caption{Sample prompt for the entailment task.}
    \label{fig:sample_prompt_entailment}
\end{figure*}

\begin{figure*}[th!]
    \centering
    \includegraphics[width=\linewidth]{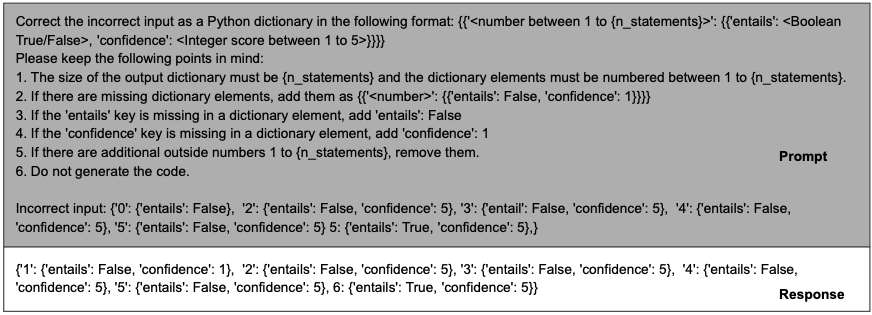}
    \caption{Sample prompt for correcting the LLM response using Llama.}
    \label{fig:sample_prompt_correction}
\end{figure*}

\end{document}